\documentclass[twoside,11pt]{article}

\usepackage{amsmath}
\usepackage{amssymb}
\usepackage{times}
\usepackage{epsfig}
\usepackage{graphicx}
\usepackage{algorithm}
\usepackage{algorithmic}
\usepackage{subfigure}

\begin{document}

\title{Robust Nonnegative Matrix Factorization via $L_1$ Norm Regularization}

\author{Bin Shen, Luo Si, Rongrong Ji, Baodi Liu\\bshen@purdue.edu}

%\author{\name Marina Meil\u{a} \email mmp@stat.washington.edu \\
%       \addr Department of Statistics\\
%       University of Washington\\
%       Seattle, WA 98195-4322, USA
%       \AND
%       \name Michael I.\ Jordan \email jordan@cs.berkeley.edu \\
%       \addr Division of Computer Science and Department of Statistics\\
%       University of California\\
%       Berkeley, CA 94720-1776, USA}
% Comment by Bin Shen

%\editor{}

\maketitle

\begin{abstract}
\begin{quote}
    Nonnegative Matrix Factorization (NMF) is a widely used technique in many applications such as face recognition, motion segmentation, etc. It approximates the nonnegative data in an original high dimensional space with a linear representation in a low dimensional space by using the product of two nonnegative matrices. In many applications data are often partially corrupted with large additive noise. When the positions of noise are known,  some existing variants  of NMF can be  applied by treating these corrupted entries as missing values. However, the positions are often unknown in many real world applications, which prevents the usage of traditional NMF or other existing variants of NMF.  This paper proposes a Robust Nonnegative Matrix Factorization (RobustNMF) algorithm that explicitly models the partial corruption as large additive noise without requiring the information of positions of noise. In practice, large additive noise can be used to model outliers. In particular, the proposed method  jointly approximates  the clean data matrix with the product of two nonnegative matrices and estimates the positions and values of outliers/noise. An efficient iterative optimization algorithm with a solid theoretical justification has been proposed to learn the desired matrix factorization. Experimental results demonstrate the advantages of the proposed algorithm.
\end{quote}
\end{abstract}

%\begin{keywords}
%    NMF
%\end{keywords}

\section{Introduction}

\noindent Nonnegative Matrix Factorization (NMF) has been widely applied in a lot of applications such as face recognition~\cite{Guillamet02}, motion segmentation~\cite{Cheriyadat09}, etc. NMF has received substantial attention due to its theoretical interpretation and practical performance.

Several variants of NMF have been proposed recently to improve the performance. Sparseness constraints have been incorporated into NMF to obtain sparse solutions \cite{Hoyer2004,Kim07}. NMF algorithms in ~\cite{Cai09,Shen2010a} are proposed to preserve the local structure on the low dimensional manifold(s). To be robust to outliers, ~\cite{Hamza06} proposes RSNMF, which is based on an outlier resistant objective function. ~\cite{Fogel07} maintains an outlier list in NMF for more robust performance.

\begin{figure}[h]
\begin{center}
\includegraphics[scale=0.3]{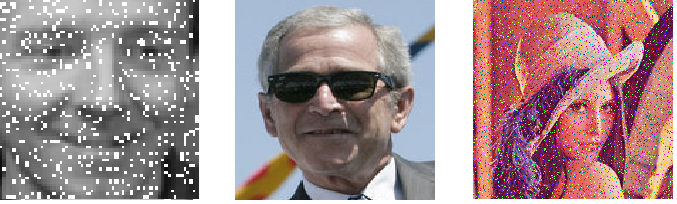}
\end{center}
\caption{\textbf{Large Additive Noise/Partial Corruption/Outlier}}
\end{figure}

In real applications, data samples are often partially corrupted(e.g, pepper and salt noise in images, occlusion on faces). Figure 1 shows some examples of this kind of partial corruption. Intuitively, partial corruption can be treated as large additive noise. Unfortunately, traditional methods based on least square estimation, such as NMF and PCA, are sensitive to this kind of noise~\cite{Torre01}, since the underlying assumption of Gaussian noise distribution is not valid. Some recent work~\cite{ShengZhang06,Eriksson10,Kaushik10} tries to deal with partial corruption. They usually assume the positions of the corruption are given ahead, and then ignore the corresponding data entries. However, it is unrealistic to assume that the positions of corruption are known in many real world applications.  ~\cite{Wright09} proposes Robust PCA to recover the noise value and position.

This paper proposes a Robust Nonnegative Matrix Factorization(RobustNMF) approach, which is able to simultaneously learn the basis matrix, coefficient matrix and estimate the positions and values of noise. The underlying observation is that the clean data allow a nonnegative factorization and the noise is sparse. An efficient iterative optimization algorithm with solid theoretical justification has been proposed to obtain the desired solution of the RobustNMF approach. To the best of our knowledge, our work is the first NMF technique that generates robust results for data wit large additive noise(partial corruption) without requiring the information of the positions of the noise.

The rest part of this paper is organized as follows. Section 2 reviews the traditional NMF algorithm. Section 3 proposes the RobustNMF algorithm, followed by the iterative optimization method in section 4. Section 5 provides some theoretical justification of the optimization method, and the experimental results are shown in section 6. Finally, we conclude and discuss future work.

\section{Review of Nonnegative Matrix Factorization}

Given a nonnegative matrix ${X} \in {\cal{R}}^{m\times n}$, each column of ${X}$ represents a data sample, the NMF algorithm aims to learn two nonnegative matrices ${U} \in {\cal{R}}^{m\times k}$ and ${V}\in {\cal{R}}^{k\times n}$  for approximating ${X}$ by the product of them, i.e. $X\approx UV$. To learn the ${U}$ and ${V}$, the following objective function should be minimized:

\begin{equation}
\begin{split}
O = ||X-UV||_{F}^{2}\\
s.t. \ \ U \ge 0, V \ge 0
\end{split}
\end{equation}

where $||.||_{F}$ denotes the Frobenius norm.

The following iterative multiplicative updating algorithm is proposed in ~\cite{Lee01} to minimize the above objective function:

\begin{equation}
U_{ij}=U_{ij}{\frac{({{X}{V}^T})_{ij}}{({U{V}{V}^T})_{ij}}}
\end{equation}

\begin{equation}
V_{ij}=V_{ij}{\frac{({{U}^T{X}})_{ij}}{({{U}^T{U}V})_{ij}}}
\end{equation}

\section{Robust Nonnegative Matrix Factorization}

The proposed Robust Nonnegative Matrix Factorization(RobustNMF) algorithm explicitly models the partial corruption, which is treated as large additive noise. Let nonnegative matrix ${X} \in {\cal{R}}^{m\times n}$ denote the observed corrupted data, while each column of ${X}$ is a data sample. Let $\hat{X} \in {\cal{R}}^{m\times n}$ denote the clean data without pollution. We have $X = \hat{X}+E$, where ${E} \in {\cal{R}}^{m\times n}$ is the large additive noise. Note that the large additive noise $E$ is not Gaussian noise with zero mean, which is well handled by least square error minimization. Moreover, we are concerned with partial corruption, and \textsl{partial} means the noise distribution is sparse. In other words, only a small portion of entries of $E$ are nonzero. For example, in face recognition, the occlusion by glasses is an instance of this kind of noise, and it covers only a small portion of the entire face.

The clean data $\hat{X}$ is approximated by $UV$(${U} \in {\cal{R}}^{m\times k}$, ${V}\in {\cal{R}}^{k\times n}$) as in traditional NMF, thus we have
\begin{equation}
\begin{split}
    X\approx UV+E
\end{split}
\end{equation}

Considering the above model and the sparseness of the large additive noise $E$, the objective function of RobustNMF is defined as follows:
\begin{equation}
\begin{split}
O_{RobustNMF} & = ||X-UV-E||_{F}^{2}\\
&+\lambda\sum_{j}[\|E_{.j}\|_0]^2\\
\end{split}
\end{equation}
The first term is to approximate the clean data; the second term is obtained from the sparseness constraint of $E$. The parameter $\lambda$ controls the tradeoff between the two terms, thus it is dependent on how large portion of entries are corrupted.

However, the $L_0$ norm in the second term makes this objective function difficult to optimize, so $L_1$ norm is employed to approximate it, which has been a popular strategy in prior research~\cite{Donoho2006}. Substituting the $L_1$ norm into the objective function, we have

%We will later discuss the side effect of this approximation.
%, even if $L_1$ norm is widely treated as a good indicator of sparseness.

\begin{equation}
\begin{split}
O_{RobustNMF} & = ||X-UV-E||_{F}^{2}\\
&+\lambda\sum_{j}[\|E_{.j}\|_1]^2\\
%& = ||X-[U,I]{V\choose E}||_{F}^{2}\\
%&+ \lambda\sum_{j}[\|E_{.j}\|_1]^2\\
& = ||X-[U,I,-I]{V\choose {E^p \choose E^n}}||_{F}^{2}\\
&+ \lambda\sum_{j}[\|E^p_{.j}\|_1+\|E^n_{.j}\|_1]^2\\
\end{split}
\end{equation}

 where $E = E^p - E^n$, $E^p = \frac{|E|+E}{2}$, $E^n = \frac{|E|-E}{2}$, and $E^p \ge 0$,$E^n \ge 0$.
 Now we have squared $L_1$ norm penalty for sparseness, which has been proved to be effective and computationally convenient~\cite{Kim07,Kim08b,Shen2010a}. Note that $E$ is the sparse large additive noise, which could be either negative or nonnegative. We need to decompose $E$ into two nonnegative matrices $E^p$ and $E^n$ described above to gain the nonnegativity which results in the convenience in optimization. We also set constraint that $X-E \ge 0$, since the clean data should be nonnegative. Finally, the objective function should be minimized with respect to $U$, $V$, $E^p$, and $E^n$ subject to the constraints that $U\ge 0$, $V\ge 0$, $E^p\ge 0$, $E^n\ge 0$, and $X-E\ge 0$.

 %The details of optimization will be described in the following section.

%-------------------------------------------------------------------------
\section{Optimization}

Since $O_{RobustNMF}$ is not convex with $U$, $V$, $E^p$, and $E^n$ jointly, it is difficult to find the global minimum for $O_{RobustNMF}$. Instead, we aim to find a local minimum  by iteratively updating $U$, $V$, $E^p$ and $E^n$ in a similar way with the work ~\cite{Lee01} for NMF.

\subsection{Update $U$}
Given $V$,$E^p$,and $E^n$, we update $U$ to decrease the value of objective function.
\begin{equation}
\begin{split}&\\
U =& arg\min_{U\ge0}||X-[U,I,-I]{V\choose {E^p \choose E^n}}||_{F}^{2}\\
&+ \lambda\sum_{j}[\|E^p_{.j}\|_1+\|E^n_{.j}\|_1]^2\\
=&arg\min_{U\ge0}||[X-E]-UV||_{F}^{2}\\
\end{split}
\end{equation}

The updating rule for $U$ to {reduce} the objective function is as follows, which can be proven in a similar way as in~\cite{Lee01}.

\begin{equation}
U_{ij}=U_{ij}{\frac{({\hat{X}{V}^T})_{ij}}{({U{V}{V}^T})_{ij}}}
\end{equation}

where $\hat{X}=X-E$. Note that at this step $E$ is given, and it satisfies the constraint that $X-E \ge 0$.%(X,0_{m\times k})$ and $\widetilde{V}=(V,\sqrt{\zeta}I_k)$.

\subsection{Update $V$, $E^p$, and $E^n$}

Now we decrease the objective function with respect to $V$, $E^p$ and $E^n$ given $U$. Let $\widetilde{V} = {V\choose {E^p \choose E^n}}$.

The updating rule for $\widetilde{V}$ is:

\begin{equation}
\begin{split}&\\
\widetilde{V}_{ij}&= max(0,\widetilde{V}_{ij} -\frac {\widetilde{V}_{ij}(\widetilde{U}^T\widetilde{U}\widetilde{V})_{ij}}{(S\widetilde{V})_{ij}}+\frac{\widetilde{V}_{ij}(\widetilde{U}^T\widetilde{X})_{ij}}{(S\widetilde{V})_{ij}})\\
\end{split}
\end{equation}

where $\widetilde{X} = {X\choose 0_{1\times n}}$, $\widetilde{U}={{U,I,-I}\choose {0_{1\times k}}{\sqrt{\lambda}e_{1\times m}}{\sqrt{\lambda}e_{1\times m}}}$, and $S$ is defined as

\begin{equation}
S_{ij} = |(\widetilde{U}^T\widetilde{U})_{ij}|
\end{equation}

%Here to insert 

\section{Correctness of Updating Rules}

To decrease the objective function with respect to $V$, $E^p$ and $E^n$. We have:

\begin{equation}
\begin{split}
(V, E^p, E^n) & = arg\min_{V,E^p,E^n\ge0} O_{RobustNMF}\\
& =arg\min_{V,E^p,E^n\ge0}||X-[U,I,-I]{V\choose {E^p \choose E^n}}||_{F}^{2}\\
&+ \lambda\sum_{j}[\|E^p_{.j}\|_1+\|E^n_{.j}\|_1]^2\\
%& =arg\min_{V,E^p,E^n\ge0}||X-[U,I,-I]{V\choose {E^p \choose E^n}}||_{F}^{2}\\
%&+ \lambda\sum_{j}[\|E^p_{.j}\|_1+\|E^n_{.j}\|_1]^2\\
& =arg\min_{V,E^p,E^n\ge0} ||{X\choose 0_{1\times n}}\\
&-{{U,I,-I}\choose {0_{1\times k}}{\sqrt{\lambda}e_{1\times m}}{\sqrt{\lambda}e_{1\times m}}}{V\choose {E^p \choose E^n}}||_{F}^{2}\\
& =arg\min_{V,E^p,E^n\ge0} ||{\widetilde{X}}-{\widetilde{U}}\widetilde{V}||_{F}^{2}\\
\end{split}
\end{equation}

where $\widetilde{X} = {X\choose 0_{1\times n}}$, $\widetilde{U}={{U,I,-I}\choose {0_{1\times k}}{\sqrt{\lambda}e_{1\times m}}{\sqrt{\lambda}e_{1\times m}}}$, $\widetilde{V} = {V\choose {E^p \choose E^n}}$.

Updating $V$,$E^p$ and $E^n$ is more involved than updating $U$, since $\widetilde{U}$ contains some negative values. Now we prove the correctness of the updating rules for $V$, $E^p$ and $E^n$ proposed in section 4.

%------------------------------------------------------------------------
%\Section{The Cool Updating Rule}

%\textbf{$W\equiv\widetilde{U}$}

%For presentation convenience, we define \textbf{$H$}, and let it hold that \textbf{$H\equiv\widetilde{V}\widetilde{\mathbf{v}}{v}$}.

\subsection{Decrease Objective Function}
\textbf{Definition 1}~\cite{Lee01} $Z(\widetilde{\mathbf{v}},\widetilde{\mathbf{v}}')$ is an auxiliary function for $F(\widetilde{\mathbf{v}})$, if it satisfies the following conditions

$Z(\widetilde{\mathbf{v}},\widetilde{\mathbf{v}}')\ge F(\widetilde{\mathbf{v}}), \ \ \ \ \ Z(\widetilde{\mathbf{v}},\widetilde{\mathbf{v}})=F(\widetilde{\mathbf{v}})$

\textbf{Lemma 1}~\cite{Lee01} If $Z$ is an auxiliary function, then $F$ is nonincreasing under the update

$\widetilde{\mathbf{v}}^{t+1} = arg\min_{\widetilde{\mathbf{v}}} Z(\widetilde{\mathbf{v}},\widetilde{\mathbf{v}}^t)$

Now we generalize the \textbf{Lemma 1} to \textbf{Lemma 2}.

\textbf{Lemma 2} If $Z$ is an auxiliary function, then $F$ is nonincreasing as long as $\widetilde{\mathbf{v}}^{t+1}$ satisfies the following condition:

$Z(\widetilde{\mathbf{v}}^{t+1},\widetilde{\mathbf{v}}^t)\le Z(\widetilde{\mathbf{v}}^{t},\widetilde{\mathbf{v}}^t)$

\textbf{Proof:}

$F(\widetilde{\mathbf{v}}^{t+1}) \le Z(\widetilde{\mathbf{v}}^{t+1},\widetilde{\mathbf{v}}^t)\le Z(\widetilde{\mathbf{v}}^{t},\widetilde{\mathbf{v}}^t) \le F(\widetilde{\mathbf{v}}^t)$ $\Box$%$\qed$

This generalization from \textbf{Lemma 1} to \textbf{Lemma 2} is similar to the generalization from EM to Generalized EM.

In our problem, $\widetilde{U}$ contains some negative value. Thus the updating rules in ~\cite{Lee01} do not hold. So we begin to seek new updating rules.

Define a matrix $S$ as follows.
\begin{equation}
S_{ij} = |(\widetilde{U}^T\widetilde{U})_{ij}|
\end{equation}

\textbf{Lemma 3} If $K(\widetilde{\mathbf{v}}^t)$ is the diagonal matrix that

\begin{equation}
K_{ab}(\widetilde{\mathbf{v}}^t) = \delta_{ab}(S\widetilde{\mathbf{v}}^t)_a/{{\widetilde{\mathbf{v}}}^t_a}
\end{equation}

then

\begin{equation}
\begin{split}&\\
Z(\widetilde{\mathbf{v}},\widetilde{\mathbf{v}}^t) =& F(\widetilde{\mathbf{v}}^t)+(\widetilde{\mathbf{v}}-\widetilde{\mathbf{v}}^t)\nabla F(\widetilde{\mathbf{v}}^t)\\
&+\frac{1}{2}(\widetilde{\mathbf{v}}-\widetilde{\mathbf{v}}^t)^TK(\widetilde{\mathbf{v}}^t)(\widetilde{\mathbf{v}}-\widetilde{\mathbf{v}}^t)
\end{split}
\end{equation}

is an auxiliary function for
\begin{equation}
F(\widetilde{\mathbf{v}}) = \frac{1}{2}\sum_i(\widetilde{\mathbf{x}}_i-\sum_a\widetilde{U}_{ia}\widetilde{\mathbf{v}}_a)^2
\end{equation}

\textbf{Proof:}

$Z(\widetilde{\mathbf{v}},\widetilde{\mathbf{v}})= F(\widetilde{\mathbf{v}})$, obviously. Now we prove that $Z(\widetilde{\mathbf{v}},\widetilde{\mathbf{v}}^t)\ge F(\widetilde{\mathbf{v}})$.

Comparing
\begin{equation}
\begin{split}&\\
F(\widetilde{\mathbf{v}}) = & F(\widetilde{\mathbf{v}}^t)+(\widetilde{\mathbf{v}}-\widetilde{\mathbf{v}}^t)\nabla F(\widetilde{\mathbf{v}}^t)\\&+\frac{1}{2}(\widetilde{\mathbf{v}}-\widetilde{\mathbf{v}}^t)^T(\widetilde{U}^T\widetilde{U})(\widetilde{\mathbf{v}}-\widetilde{\mathbf{v}}^t)
\end{split}
\end{equation}

to the $Z(\widetilde{\mathbf{v}},\widetilde{\mathbf{v}}^t)$, we find that we only need to show
\begin{equation}
\begin{split}&
(\widetilde{\mathbf{v}}-\widetilde{\mathbf{v}}^t)^TK(\widetilde{\mathbf{v}}^t)(\widetilde{\mathbf{v}}-\widetilde{\mathbf{v}}^t)-(\widetilde{\mathbf{v}}-\widetilde{\mathbf{v}}^t)^T(\widetilde{U}^T\widetilde{U})(\widetilde{\mathbf{v}}-\widetilde{\mathbf{v}}^t) \ge 0\\
&(\widetilde{\mathbf{v}}-\widetilde{\mathbf{v}}^t)^T[K(\widetilde{\mathbf{v}}^t)-\widetilde{U}^T\widetilde{U}](\widetilde{\mathbf{v}}-\widetilde{\mathbf{v}}^t) \ge 0
\end{split}
\end{equation}

To prove the positive semidefiniteness, consider the matrix $M(\widetilde{\mathbf{v}}^t)$:

\begin{equation}
M_{ab}(\widetilde{\mathbf{v}}^t) = \widetilde{\mathbf{v}}^t_a[K(\widetilde{\mathbf{v}}^t)-\widetilde{U}^T\widetilde{U}]_{ab}\widetilde{\mathbf{v}}^t_b
\end{equation}

$M$ is a rescaling of $K(\widetilde{\mathbf{v}}^t)-\widetilde{U}^T\widetilde{U}$. The $M$ is semipositive definite if and only if $K(\widetilde{\mathbf{v}}^t)-\widetilde{U}^T\widetilde{U}$ is.
\begin{equation}
\begin{split}&\\
\mu^T M \mu &= \sum_{ab}\mu_a M_{ab} \mu_b\\
& = \sum_{ab}\widetilde{\mathbf{v}}^t_aS_{ab}\widetilde{\mathbf{v}}^t_b\mu^2_a-\mu_a \widetilde{\mathbf{v}}^t_a (\widetilde{U}^T\widetilde{U})_{ab}\widetilde{\mathbf{v}}^t_b \mu_b\\
& = \sum_{ab}S_{ab}\widetilde{\mathbf{v}}^t_a\widetilde{\mathbf{v}}^t_b[\frac{1}{2}\mu_a^2+\frac{1}{2}\mu_b^2-sgn((\widetilde{U}^T\widetilde{U})_{ab})\mu_a\mu_b]\\
& = \sum_{ab}S_{ab}\widetilde{\mathbf{v}}^t_a\widetilde{\mathbf{v}}^t_b\frac{1}{2}[\mu_a-sgn((\widetilde{U}^T\widetilde{U})_{ab})\mu_b]^2\\
& \ge 0
\end{split}
\end{equation}

where
\begin{equation}
%\begin{split}
sgn(x)=\left\{
\begin{array}{r@{:\quad}l}
-1& x <0 \\0 & x = 0 \\ 1 & x >0
\end{array}
\right.
%\end{split}
\end{equation}

Note in our setting, $\widetilde{U}$ contains some negative values, but $\widetilde{V}$ is nonnegative. $\Box$

Substitute \textbf{Lemma 3} into \textbf{Lemma 1}, the updating rule is:
\begin{equation}
\begin{split}&\\
\widetilde{\mathbf{v}}^{t+1}&= \widetilde{\mathbf{v}}^t - K(\widetilde{\mathbf{v}}^t)^{-1}\nabla F(\widetilde{\mathbf{v}}^t)\\
&=\widetilde{\mathbf{v}}^t - K(\widetilde{\mathbf{v}}^t)^{-1}\widetilde{U}^T\widetilde{U}\widetilde{\mathbf{v}}^t+K(\widetilde{\mathbf{v}}^t)^{-1}\widetilde{U}^T\widetilde{\mathbf{x}}\\
\end{split}
\end{equation}

Writing the components explicitly, we get:
\begin{equation}
\begin{split}&\\
\widetilde{\mathbf{v}}^{t+1}_a&= \widetilde{\mathbf{v}}^t_a -\frac {\widetilde{\mathbf{v}}^t_a(\widetilde{U}^T\widetilde{U}\widetilde{\mathbf{v}}^t)_a}{(S\widetilde{\mathbf{v}}^t)_a}+\frac{\widetilde{\mathbf{v}}^t_a(\widetilde{U}^T\widetilde{\mathbf{x}})_a}{(S\widetilde{\mathbf{v}}^t)_a}\\
%&=h^t - K(h^t)^{-1}W^TWh^t+K(h^t)^{-1}W^Tx\\
\end{split}
\end{equation}

The proposed updating rules can deal with negative values by explicitly considering the negative part of large additive noise in the $\widetilde{U}$. If $\widetilde{U} \ge 0$, the first two terms in the above updating rule would cancel each other, resulting in the same rule as in ~\cite{Lee01}.

The $\widetilde{V}$ gained by (22) is made up of three parts: $V$,$E^p$,and $E^n$. All of them should be nonnegative. Unfortunately, the value $\widetilde{\mathbf{v}}^{t+1}$ gained by the rule (22) does not guarantee the nonnegativity. Now we discuss how to keep it nonnegative while updating the values.

In the auxiliary function $Z$, we see that the $K(\widetilde{\mathbf{v}}^t)$ is a diagonal matrix. Thus the second order terms only involve the form $\widetilde{\mathbf{v}}_a^2$. This results in a very important property in \textbf{Lemma 4}.

\textbf{Lemma 4} If ${\widetilde{\mathbf{v}}}^{t+1} = arg\min_{\widetilde{\mathbf{v}}} Z(\widetilde{\mathbf{v}},{\widetilde{\mathbf{v}}}^t)$ and $\widetilde{\mathbf{v}}^{t}\ge \widetilde{\mathbf{v}}'^{t+1} \ge {\widetilde{\mathbf{v}}}^{t+1}$, then $Z({\widetilde{\mathbf{v}}'^{t+1}}, {\widetilde{\mathbf{v}}}^t) \le Z(\widetilde{\mathbf{v}}^t,\widetilde{\mathbf{v}}^t)$.

%\begin{figure}[h]
%\begin{center}
%\includegraphics[scale=0.3]{1.png}
%\end{center}
%\caption{\textbf{Lemma 4}}
%\end{figure}

According to \textbf{Lemma 2} and  \textbf{Lemma 4}, we have $F(\widetilde{\mathbf{v}}'^{t+1})\le F(\widetilde{\mathbf{v}}^{t})$.

So, to ensure the nonnegativity, we can simply threshold $\widetilde{\mathbf{v}}^{t+1}$ by $0$. This operation will introduce the nonnegativity, while keeping the value of $F$ nonincrease from $\widetilde{\mathbf{v}}^{t}$ to the thresholded $\widetilde{\mathbf{v}}^{t+1}$. Thus, we have the updating rule described in (9).

%-------------------------------------------------------------------------
\subsection{Convergence Analysis}

Since the objective function has a lower bound, e.g., 0, and the updating rules for $U$, $V$, and $E$ will all cause the objective function nonincrease, the algorithm always converges.

\section{Experimental Results}

This section presents experimental results on two different applications of noise detection(identifying exact positions of large additive noise), and image reconstruction/denoising.

\subsection{Large Additive Noise Detection}

Several algorithms have been proposed to deal with data with partial corruption. However, they usually assume the positions of noise are not known in advance. Fortunately, the proposed RobustNMF is able to locate the positions. Once the missing values are located, existing algorithms can also be applied. This subsection presents experiments for detecting the positions of large noise in face images and image patches. The reported results are averaged over ten runs.
%\subsubsection{Large Additive Noise Detection on Face Images}

The experiments are based on the ORL face dataset. Each face image is of size 32$\times$32, thus is represented by a 1024 dimensional vector. For each face in a randomly selected subset, 50 pixels are randomly selected and replaced with the values of 255 to simulate the large additive noise. The polluted faces make up the data matrix $X$, each column of which corresponds to a polluted face image. Then, we apply the RobustNMF algorithm to this $X$ to estimate $U$, $V$ and $E$. Furthermore, we scan all the entries of the $E$. When $E_{ij}$ is nonzero, we claim that the corresponding pixel is polluted. With the above procedure, we are able to detect the positions of noise by analyzing $E$. The performance is evaluated by precision and recall, where $Precision = \frac{\#Detected Polluted Pixels}{\#Nonzero Entries In E}\times100\%$ and $Recall = \frac{\#Detected Polluted Pixels}{\#Total Pixels Polluted}\times100\%$. Here we only show the performance of our algorithm, since no other algorithm, to the best of our knowledge, is designed to handle this task. We tried to compare the proposed algorithm with PCA and NMF. We first applied PCA  or NMF to the noisy data $X$ to gain a reconstruction $\hat{X}$, and then tried to detect the positions of noise by analyzing the difference $X-\hat{X}$. However, it is very difficult to find an appropriate threshold of the difference and the performance is very sensitive to this. We tried several thresholds, all of which gave poor results, probably because the partial corruption significantly skews the solution of PCA and NMF.
In figure 2, the left subfigure presents the precision and recall versus different numbers of face images. This algorithm gains a precision of over 90\%, and a recall of over 50\%. The performance increases with the increase of the number of image faces. This is reasonable, since more samples means there is more information that RobustNMF can explore. When the number is large enough, increasing the number does not help to improve the performance any longer. Here $k$ is set to 10, and $\lambda$ is set to 0.04.

\begin{figure*}
\begin{center}
 % \centering
  \includegraphics[width = 6.2in, height = 1.6in]{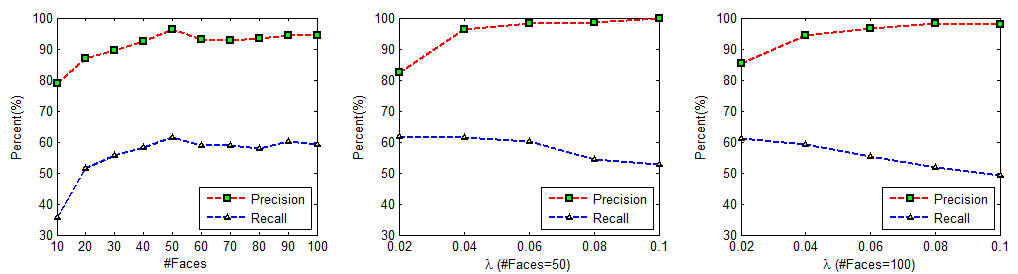}%6 1.3
 \end{center}
  \caption{Noise Detection Results in Face Images}
  \label{fig_sample}
\end{figure*}

%\caption{Noise Detection Results in Face Images - Left:\textbf{Precision, Recall V.S. \#Faces} Middle:\textbf{Precision, Recall V.S. $\lambda$ (\#Faces = 50)} Right:\textbf{Precision, Recall V.S. $\lambda$ (\#Faces = 100)}}

  The middle and right subfigures in figure 2 investigate the relationship between performance and the parameter $\lambda$. The $k$ is still set to 10, and the number of face images is fixed at 50 in middle subfigure, and 100 in right subfigure. Generally speaking, the algorithm gains over 90\% precision, and over 50\% recall. With larger values of $\lambda$, the precision will become a little higher, and the recall a little lower. This is consistent with our expectation, since a larger $\lambda$ indicates the detected noise is more sparse, which often leads to higher precision and lower recall.

\subsection{Image Reconstruction/Denoise}
 This subsection presents the performance of RobustNMF on reconstruction/denoising. First, we simulate large additive noise in the same way as in previous subsection. Given the polluted data $X$, RobustNMF will learn $U$, $V$ and $E$. The original images are reconstructed as $UV$.

As discussed before, some other algorithms, such as WNMF~\cite{ShengZhang06}, are able to handle large noise if the positions are given. Thus, we can use RobustNMF to locate noise and then employ WNMF to recover the images. The reason for combining these two methods is that there is an approximation of $L_0$ norm by $L_1$ in RobustNMF, which may cause the the absolute value of estimation of $E$ smaller than the truth. Let RobustNMF+WNMF denote the new combined method.

\subsubsection{Reconstruction of Faces}

A subset of faces from ORL face dataset are selected, and same large noise is added to generate a set of polluted samples denoted as $X$ in a similar way as described in Section 6.1, while the original data samples form the matrix $\widetilde{X}$. Mean Squared Error(MSRE) is used to measure the reconstruction performance.

For NMF, matrices $U$ and $V$ are learned based on $X$, and then are used in reconstruction. Compared with the original noise free matrix $\widetilde{X}$, the MSRE is calculated as ${\frac{1}{N}}||\widetilde{X}-UV||_F^2$, where $N$ is the number of samples. For RobustNMF, the MSRE definition is the same as for NMF. For RobustNMF+WNMF, based on $X$, RobustNMF learns $U$, $V$, and $E$. Since $E$ is an indictor for whether a pixel is polluted or not. Taking $E$ as a mask, WNMF learns the new matrix $\widetilde{U},\widetilde{V}$. The MSRE is defined as ${\frac{1}{N}}||\widetilde{X}-\widetilde{U}\widetilde{V}||_F^2$.

Experiments are conducted with varying number of pixels polluted and faces. In the first set of experiments,we fix the number of faces to be 50 or 100, and then vary the number of pixels polluted, from 10 to 100 with a step of 10. This means that there are about 1 percent to 10 percent pixels corrupted in each face. The results are shown in top row of figure 3. In the second set of experiments, the number of pixels polluted is fixed at 50 or  100, which means about 5 or 10 percent of pixels on each face are corrupted. Experiments are conducted with various numbers of faces, from 10 to 100 with a step of 10. The results are shown in bottom row of figure 3.

It can be seen from these experiments that both RobutNMF and RobustNMF+WNMF consistently outperform the traditional NMF with varying number of data samples. With the increasing amount of noise, the advantages of proposed algorithms become even larger. This is because RobustNMF is able to detect the positions of the large value noises, i.e. the partial corruption, which enables the application of WNMF. Considering the approximation of $L_0$ norm by $L_1$ norm, the large noise is underestimated, and that is why we prefer Robust+WNMF to pure RobustNMF, even though both methods outperform the traditional NMF.

%\begin{figure*}
%\begin{center}
% % \centering
%  \includegraphics[scale=0.45]{6_m_2.png}
% \end{center}
%  \caption{Face Reconstruction Results - Left: MSRE V.S. \# Faces; Right: MSRE V.S. \# Polluted Pixels in Each Face}
%  \label{fig_sample}
%\end{figure*}
\begin{figure*}
\begin{center}
 % \centering
  \includegraphics[width=5.55in,height=3.0in]{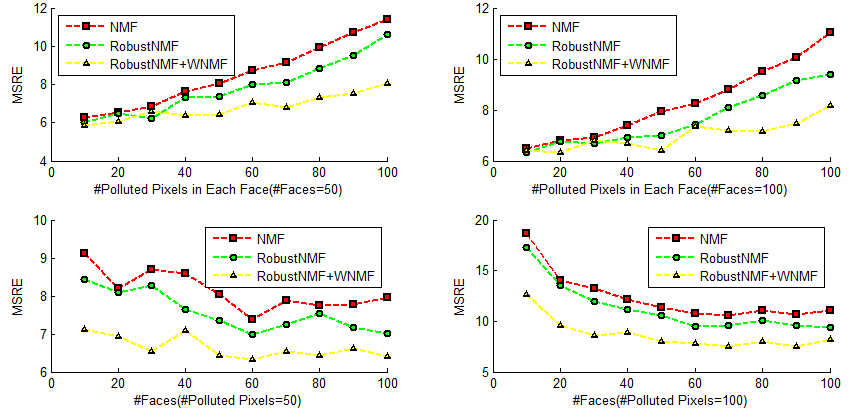}%3.5 2.1%[scale=0.4] [width=1.5in,height=0.6in] [width=1.2\linewidth]
 \end{center}
  \caption{Face Reconstruction Results - Top: MSRE V.S. \# Polluted Pixels in Each Face; Bottom: MSRE V.S. \# Faces}
  \label{fig_sample}
\end{figure*}
%\begin{figure*}
%\begin{center}
% % \centering
%  \includegraphics[scale=0.45]{6_m_4.png}
% \end{center}
%  \caption{Face Reconstruction Results - Left: MSRE V.S. \# Faces; Right: MSRE V.S. \# Polluted Pixels in Each Face}
%  \label{fig_sample}
%\end{figure*}

\subsubsection{Image Denoising(Reconstruction of Patches)}

This subsection presents some experiment results on image denoising by using RobustNMF. Pepper and salt noise is added to natural images. The noise density is set to 5\%, which means about 5\% of pixels are affected. The noisy image is converted into a set of patches, to which RobustNMF is applied. $\lambda$ is set to 0.04 and $k$ is set to 10. $UV$ is used to reconstruct the original image. Some denoising results are shown in figure 4. The first row shows the generated polluted images, the second row shows the denoised results by traditional NMF, and the third row is the results by RobustNMF. It can be seen that RobustNMF outperforms traditional NMF. Due to the space limit, the ground truth image, the results by RobustNMF+WNMF, and more experiments on other images are given in supplemental materials.

\begin{figure*}
\begin{center}
 % \centering
  \includegraphics[scale = 0.49]{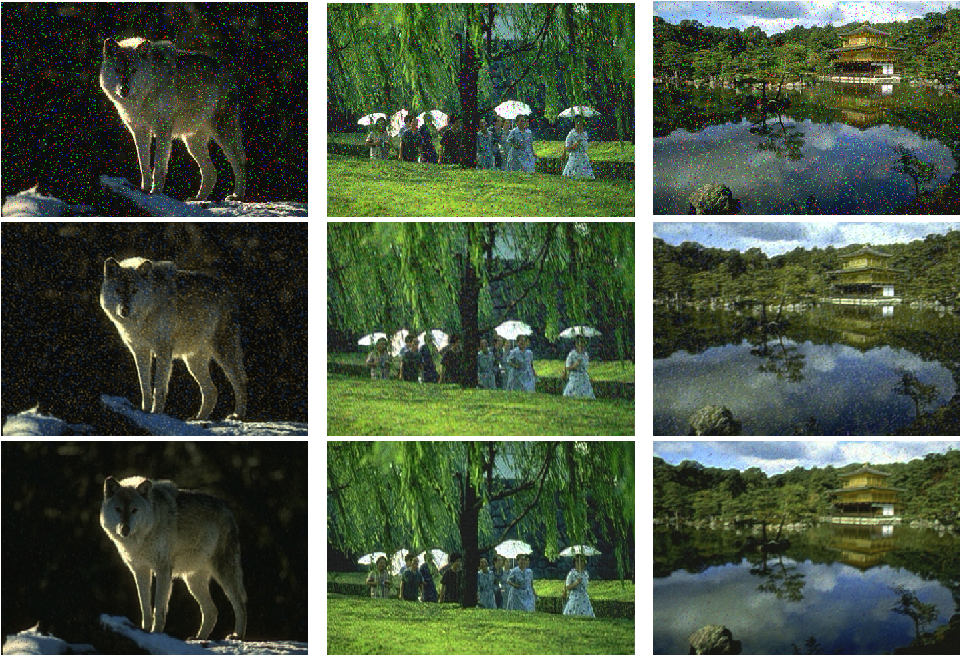}%scale=0.52 width=6in,height=3.5in
 \end{center}
  \caption{Image Denoising Results - Top: Polluted Images; Middle: Results by NMF; Bottom: Results by RobustNMF}
  \label{fig_sample}
\end{figure*}

\section{Conclusion}
Data in many real world applications are often partially corrupted without the explicit information of positions of noise, which prevents the usage of NMF and other existing variants.This paper proposes a RobustNMF algorithm for large additive noise, which can handle partial corruption without requiring the position information of noise in advance. The proposed algorithm is able to simultaneously locate and estimate the large additive noise and learn the basis matrix $U$ and coefficient matrix $V$ in the framework of NMF. This proposed algorithm also paves the way to apply other variants of NMF(e.g. WNMF) to data with missing values by estimating the positions of noise. An efficient optimization algorithm with a solid theoretical justification is proposed for RobustNMF. Experimental results on three different sets of applications demonstrate the advantages of our algorithm.

As for future research, we plan to explore a low rank version of RobustNMF, which can automatically find the adequate low rank of the decomposed matrices. Similar to RobustPCA~\cite{Wright09}, more applications of RobustNMF can be investigated, since NMF is widely used in various areas, including computer vision, text mining, speech analysis, and etc.

%\vskip 0.2in
\bibliographystyle{IEEEbib}
\bibliography{RobustNMFarch}

%\begin{thebibliography}{10}
%\itemsep=-3.pt
%\small
%\footnotesize

%\bibitem{hsieh11}
%C.-J. Hsieh, Inderjit S. Dhillon.
%\newblock Fast Coordinate Descent Methods with Variable Selection for Non-negative Matrix Factorization.
%\newblock {\em KDD}, 2011.
%
%\bibitem{Lee01}
%Daniel D. Lee, H. Sebastian Seung
%\newblock Algorithms for Non-negative Matrix Factorization.
%\newblock {\em NIPS}, 2001.
%
%\end{thebibliography}

\end{document}